\documentclass[11pt,a4paper]{article}
\usepackage[hyperref]{acl2019}
\usepackage{times}
\usepackage{latexsym}

\usepackage{url}
\usepackage{booktabs}       
\usepackage{amsfonts}       
\usepackage{nicefrac}       
\usepackage{microtype}      
\usepackage{graphicx}
\usepackage{capt-of}
\usepackage{float}

\usepackage{url}
\usepackage{amsfonts, amssymb}       
\usepackage{nicefrac}       
\usepackage{microtype}      
\usepackage{graphicx}
\usepackage{capt-of}
\usepackage{float}
\usepackage{macros}
\usepackage{mathtools}
\usepackage{array}
\usepackage{url}
\aclfinalcopy 

\title{Deep Reinforcement Learning For Modeling Chit-Chat Dialog With Discrete Attributes}

\author{Chinnadhurai Sankar \thanks{~~Work done during internship at Google}\\
  Mila, Universit\'e de Montr\'eal \\
  \texttt{chinnadhurai@gmail.com} \\\And
  Sujith Ravi \\
  Google Research \\
  \texttt{sravi@google.com} \\}

\date{}

\begin{document}
\maketitle
\begin{abstract}
Open domain dialog systems face the challenge of being repetitive and producing generic responses. In this paper, we demonstrate that by conditioning the response generation on {\it interpretable} discrete dialog attributes and {\it composed} attributes, it helps improve the  model perplexity and results in diverse and interesting non-redundant responses.
We propose to formulate the dialog attribute prediction as a reinforcement learning (RL) problem and use policy gradients methods to optimize utterance generation using long-term rewards. Unlike existing RL approaches which formulate the token prediction as a policy, our method reduces the complexity of the policy optimization by limiting the action space to dialog attributes, thereby making the policy optimization more practical and sample efficient. We demonstrate this with experimental and human evaluations.
\end{abstract}

\section{Introduction}
Following the success of neural machine translation systems \citep{bahdanau2014neural,seq2seq,cho_NMT}, there has been a growing interest in adapting the encoder-decoder models to model open-domain conversations \citep{sordoniDialog,hred,latentHRED,vinyals2015neural}.This is done by framing the next utterance generation as a machine translation problem by treating the dialog history as the source sequence and the next utterance as the target sequence. Then the models are trained end-to-end with Maximum Likelihood (MLE) objective without any hand crafted structures like slot-value pairs, dialog manager, etc used in conventional dialog modeling \citep{LagusK02_dialog_topic_tracking}.
Such data driven approaches are worth pursuing in the context of open-domain conversations since the next utterance distribution in open-domain conversations exhibit high entropy which makes it impractical to manually craft good features.

While the encoder-decoder approaches are promising, lack of specificity has been one of the many challenges \citep{genericResponse} in modelling non-goal oriented dialogs. Recent encoder-decoder based models usually tend to generate generic or dull responses like ``\textit{I don't know.}". One of the main causes are the implicit imbalances present in the dialog datasets that tend to potentially handicap the models into generating uninteresting responses.

Imbalances in a dialog dataset can be broadly divided into two categories: \textit{many-to-one} and \textit{one-to-many}. \textit{Many-to-one} imbalance occurs when the dataset contain very similar responses to several different dialog contexts. In such scenarios, decoder learns to ignore the context (considering it as noise) and behaves like a regular language model. Such a decoder would not generalize to new contexts and will end up predicting generic responses for all contexts. In the \textit{one-to-many} case, the dataset may exhibit a different type of imbalance where a certain type of generic response may be present in abundance compared to other plausible interesting responses for the same dialog context \cite{genericResponse}. 
When trained with a maximum-likelihood (MLE) objective, generative models usually tend to place more probability mass around the most commonly observed responses for a given context. So, we end up observing little variance in the generated responses in such cases. While these two imbalances are problematic for training a dialog model, they are also inherent characteristics of a dialog dataset which cannot be removed.

Several approaches have been proposed in the literature to address the generic response generation issue. \citet{li2015diversity} propose to modify the loss function to increase the diversity in the generated responses. Multi-resolution RNN \citep{mrRNN} addresses the above issue by additionally conditioning with entity information in the previous utterances. Alternatively, \citet{hybridRetrieval} uses external knowledge from a retrieval model to condition the response generation. Latent variable models inspired by Conditional Variational Autoencoders (CVAEs) 
are explored in \citep{conditionalVAEGen,zhou_VAE_dialog}. While models with continuous latent variables tend to be uninterpretable, discrete latent variable models exhibit high variance during inference. \citet{conditionalVAEGen} append discrete attributes such as sentiment to the latent representation to generate next utterance. 

\subsection{Contributions} 

{\bf New Conditional Dialog Generation Model.} Drawing insights from \citep{conditionalVAEGen,emotion_chatting_machine}, we propose a {\it conditional utterance generation model} in which the next utterance is conditioned on the dialog attributes corresponding to the next utterance. To do this, we first predict the higher level dialog attributes corresponding to the next response. Then we generate the next utterance conditioned on the dialog context and predicted attributes. Dialog attribute of an utterance refers to discrete features or aspects associated with the utterance. Example attributes include dialog-acts, sentiment, emotion, speaker id, speaker personality or other user defined discrete features of an utterance. While previous research works lack the framework to learn to predict the attributes of the next utterance and mainly view the next utterance's attribute as a control variable in their models, our method learns to predict the attributes in an end-to-end manner. This alleviates the need to have utterances annotated with attributes during inference.

{\bf RL for Dialog Attribute Selection.}
Further, it also enables us to formulate the dialog attribute selection as a reinforcement learning (RL) problem and optimize the policy initialized by the supervised training using REINFORCE \citep{reinforce}. While the Supervised pre-training helps the model to generate utterances coherent with the dialog history, the RL formulation encourages the model to generate utterances optimized for long term rewards like diversity, user-satisfaction scores etc. This way of optimizing the policy over the discrete dialog attribute space is more practical as the action space is low dimensional instead of the entire vocabulary (as common in policies which involve predicting the next token to generate). 

By using REINFORCE \citep{reinforce} to further optimize the dialog attribute selection process, We then show improvements in specificity of the generated responses both qualitatively (based on human evaluations) and quantitatively (with respect to the \textit{diversity} measures). The diversity scores, \textit{distinct-1} and \textit{distinct-2} are computed as the fraction of uni-grams and bi-grams in the generated responses as described in \citep{li2015diversity}.

{\it Improvements on Dialog datasets demonstrated through quantitative \& qualitative Evaluations:} Additionally, we annotate an existing open domain dialog dataset using dialog attribute classifiers trained with tagged datasets like Switchboard \citep{switchboard,jurafsky_switchboard}, Frames \citep{frames} and demonstrate both quantitative (in terms of token perplexity/embedding metrics \citep{rus2012comparison_embedding_metrics,mitchell2008vector_embedding_metrics}) and qualitative improvements (based on human evaluations) in generating interesting responses. In this work, we show results with two types of dialog attributes - sentiment and dialog-acts.
It is worth investigating this approach as we need not invest much in training classifiers for very high accuracy and we show empirically that annotations from classifiers with low accuracy are able to boost token perplexity. 
We conjecture that the irregularities in the auto-annotated dialog attributes induce a regularization effect while training deep neural networks analogous to the dropout mechanism. Also, annotating utterances with many types of dialog attributes could increase the regularization effect and potentially tip the utterance generation in the favor of certain low frequency but interesting responses.

In this work, we are mainly interested in exploring the impact of the jointly modelling extra discrete dialog attributes along with dialog history for next utterance generation and their contribution to addressing the generic response problem. Although our approach is flexible enough to include latent variables additionally, we mainly focus on the contribution of dialog attributes to address the "generic" response issue in this work.
\section{Attribute Conditional HRED}

In this paper, we extend the \textit{HRED} \citep{hred} model (elaborated in the Appendix section) by jointly modelling the utterances with the dialog attributes of each utterance.
\textit{HRED} is a encoder-decoder model consisting of a token-level RNN encoder and an utterance-level RNN encoder to summarize the dialog context followed by a token-level RNN decoder to generate the next utterance.
The joint probability can be factorized into dialog attributes prediction, followed by next utterance generation conditioned on the predicted dialog attributes as shown in equation \ref{eq:conditional_generataion}
.
\begin{multline}
\tiny
\label{eq:conditional_generataion}
P(\mathrm{U_{m}}, \mathrm{DA_{1:K}}|\mathrm{U_{1:m-1}}) \\ = \prod_{i=1}^{K}P(\mathrm{DA_{i}}|\mathrm{U_{1:m-1}}) * P(\mathrm{U_{m}}|\mathrm{U_{1:m-1}}, \mathrm{DA_{1:K}})
\end{multline}
where $\mathrm{DA_{1:K}}$ denote $K$ different dialog attributes corresponding to the utterance $\mathrm{U_{m}}$. $\mathrm{U_m}$ is the $m_{th}$ utterance, $\mathrm{U_{1:m-1}}$ are the past utterances. For instance, if we condition on three dialog attributes - \textit{sentiment, dialog-acts and emotion}, we would have $K=3$. Further, we assume that the dialog attributes are conditionally independent given the dialog context. More simply, we predict the attributes of the next utterance and then, condition on the previous context \& the predicted attributes to generate the next utterance.
\begin{figure}[htbp]

\centering
    \includegraphics[scale=0.22]{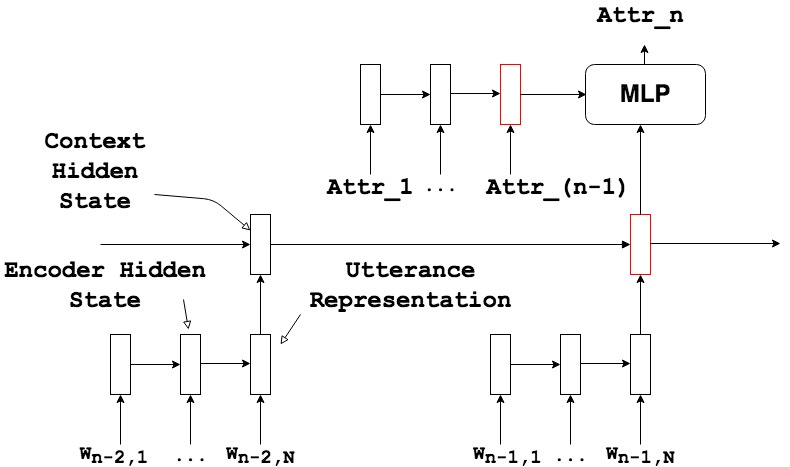}
    \caption{Dialog attribute classification: We predict the dialog attribute of the next utterance based on the previous context and attributes corresponding to the previous utterances. Please note that we depict only a single attribute for convenience}
    \label{fig:dialog2act}
    \end{figure}

\begin{figure}
\centering
    \includegraphics[scale=0.25]{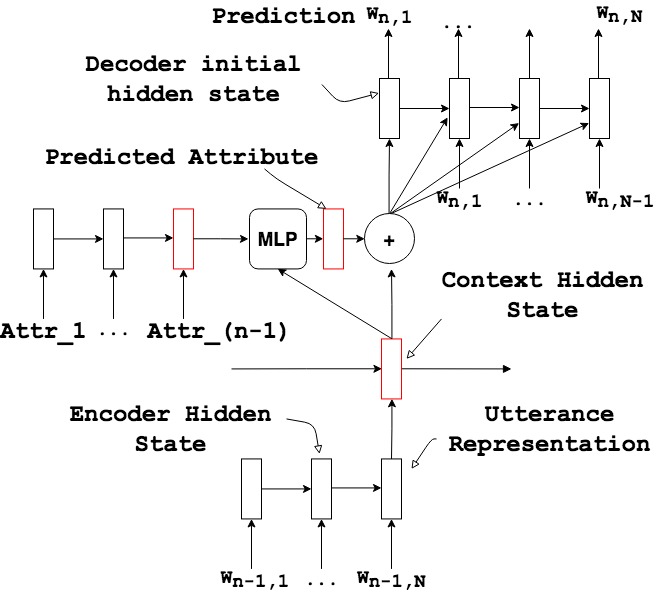}
    \caption{Attribute Conditional HRED : Token generation is additionally conditioned on the predicted dialog attributes. The dialog attribute's embedding is concatenated with the context vector. }
    \label{fig:utr2act2utr}

\end{figure}

\subsection{Dialog Attribute Prediction}
We predict the dialog attribute of the next utterance conditioned on the context vector i.e. summary of the previous utterances and the dialog attributes of the previous utterances. We first pass the attributes of all the previous utterances through an RNN. We combine only the last hidden state of this RNN with the context vector (represents the summary of all the previous utterances) to predict the dialog attribute of the next utterance as shown in Figure \ref{fig:dialog2act}.

If the dialog dataset is not annotated with the dialog attributes, we build a classifier (with a manually tagged dataset) to annotate the dialog attributes. This classifier is a simple MLP. We empirically show that this classifier need not have high accuracy to improve the dialog modeling. We hypothesize that few misclassified attributes could potentially provide a regularization effect similar to the dropout mechanism \citep{srivastava2014dropout}.

\subsection{Conditional Response Generation}
After the dialog attributes prediction, we generate the next utterance conditioned on the dialog context and the predicted attributes as shown in Figure \ref{fig:utr2act2utr}. Token generation of the next utterance is modelled as in equation \ref{eq:conditional_token_generataion}. The context and attributes are combined by concatenating their corresponding hidden states.
\begin{equation}
\large
\label{eq:conditional_token_generataion}
h_{dec_{m,n}} = \mathrm{f_{dec}}(h_{dec_{m,n-1}}, w_{m,n-1}, \mathbf{c_{m}}) 
\end{equation}
where $h_{dec_{m,n}}$ is the recurrent hidden state of the decoder after seeing $n-1$ words in the $m$-th utterance, $\mathrm{f_{dec}}$ is the token level response decoder, and
\begin{equation}
\large
\mathbf{c_{m}}=[s_{m-1}; da^{1}_{m}; da^{2}_{m}; ...; da^{K}_{m}]    
\end{equation}
where $s_{m-1}$ is the summary of previous $m-1$ utterances (recurrent hidden state of the utterance-level encoder), and $da^{1}_{m}, da^{2}_{m}, ..., da^{K}_{m}$ are the $K$ dialog attribute embeddings corresponding to the $m$-th utterance.  

During inference, we first predict the dialog attributes of the dialog context. We then predict the dialog attribute of the next utterance conditioned on the predicted attribute and the hierarchical utterance representations. We combine the predicted attribute's embedding vector with the context representation to generate the next utterance. 
Looking from another perspective, we could formulate the conditional utterance generation problem as a multi-task problem where we jointly learn to predict the dialog attributes and tokens of the next utterance.

\subsection{RL for Dialog Attribute Prediction} \label{sec: RL for Dialog Attribute Prediction}
Often the MLE objective does not capture the true goal of the conversation and lacks the framework which can take developer-defined rewards into account for modelling such goals. Also, the MLE-based seq2seq models fail to model long term influence of the utterances on the dialog flow causing coherency issues. This calls for a Reinforcement Learning (RL) based framework which has the ability to optimize policies for maximizing long term rewards. At the core, the MLE objective tries to increase the conditional utterance probabilities and influences the model to place higher probabilities over the commonly occurring utterances. On the other hand, RL based methods circumvent this issue by shifting the optimization problem to maximizing long term rewards which could promote diversity, coherency, etc.

Previous approaches \citet{jiweili2016_rl,naturalLAnguageDoesNotEmerge,dealOrNoDeal_2017} propose to model the token prediction of the next utterance as a reinforcement learning problem and optimize the models to maximize hand-crafted rewards for improving diversity, coherency, and ease of answering.
Their approaches involves pre-training the encoder-decoder models with supervised training and then refining the utterance generation further with RL using the hand-engineered rewards.
Their state space consists of the dialog context representation (encoder hidden states). 
Their action space at a given time step includes all possible words that the decoder can generate (which is very large).

While this approach is appealing, policy gradient methods are known to suffer from high variance when using large action spaces. This makes training extremely unstable and requires significant engineering efforts to train successfully.

Another potential drawback with directly acting over the vocabulary space is that the RL optimization procedure tends to strip away the linguistic / natural language aspects learned during the supervised pre-training step, as observed in \citep{naturalLAnguageDoesNotEmerge,dealOrNoDeal_2017}. Since the primary focus of the RL objective function is to improve the final reward (which may not emphasize on the linguistic aspects of the generated responses, for e.g., diversity scores), the optimization algorithm could lead the decoder into generating unnatural responses.
We propose to avoid both the issues by reducing the action space to a higher level abstraction space i.e. the dialog attributes. 
Our action space comprises the discrete dialog attributes and the state space is the dialog context.
Intuitively, this enables the RL policy to view the dialog attributes as control variables for improving dialog flow and modelling long term influence. 
For instance, if the input response was ``\textit{how old are you?}", an RL policy optimized to maximize conversation length and engagement could choose to set one of the next utterance attributes as a question-type to generate a response like ``\textit{why do you ask?}" instead of a straightforward answer, to keep the conversation engaging. 
Thus, we believe that this approach enables the model to predict such rare but interesting utterances to which the MLE objective fails to give attention. 

Our policy network comprises of the encoders and the attribute prediction network. Given the previous utterances $\mathrm{U_{1:m-1}}$, the policy network first encodes them by using the encoders. Then this encoded representation is passed to the attribute prediction network. The output of the attribute prediction network is the action. While there are many ways to design the reward function, we adopt the \textit{ease-of-answering} reward introduced by \citet{jiweili2016_rl} - negative log-likelihood of a set of manually constructed dull utterances (usually the most commonly occurring phrases in the dataset) in response to the next generated utterance. Let $\mathbb{S}$ be the set of dull utterances. With the sampled dialog-acts, $\mathrm{DA_{1:K}}$ from the policy network, we generate the next utterance $\mathrm{U_{m}}$ using the decoder. Then we add this generated utterance to the context and predict the probability of seeing one of the dull utterances in the $m+1$-th step. This is used to compute the reward as follows:
\begin{align}
\label{reward}
R = \frac{1}{|{\mathbb{S}}|} \sum_{s \in \mathbb{S}} \frac{1}{N_{s}} \mathrm{log} P(s|\mathrm{U_{1:m}}),
\end{align}
where $N_{s}$ is the number of tokens in the dull utterance $s$. The normalization avoids the reward function attending to only the longer dull responses. We use REINFORCE \citep{reinforce} to optimize our policy, $P_{RL}(\mathrm{DA_{1:K}}|\mathrm{U_{1:m-1}})$. 
The expected reward is given by equation \ref{eq:expected_reward}. 
\begin{equation}
\label{eq:expected_reward}
J(\theta) = \mathbb{E}[R(\mathrm{U_{1:m-1}}, \mathrm{DA_{1:K}})]
\end{equation}
The gradient is estimated as in equation \ref{eq:gradient_expected_reward}. 
\begin{equation}
\label{eq:gradient_expected_reward}
\nabla J(\theta_{RL}) = (R - b) \nabla\mathrm{log}P_{RL}(\mathrm{DA_{1:K}}|\mathrm{U_{1:m-1}}),
\end{equation}

where $b$ is the  reward baseline (computed as the running average of the rewards during training). We initialize the policy with the supervised training and add an L2-loss to penalize the network weights from moving away from the supervised network weights. \section{Training Setup}
\textbf{Datasets}: We first start with the Reddit-discourse dataset \citep{reddit_cleaned_google} for training dialog attribute classifiers and modelling utterance generation.\\
\textit{Reddit}: The Reddit discourse dataset \citep{reddit_cleaned_google} is manually pre-annotated with dialog-acts via crowd sourcing.
The dialog-acts comprise of \textit{answer, question, humor, agreement, disagreement, appreciation, negative reaction, elaboration, announcement}.
It comprises conversations from around $9000$ randomly sampled Reddit threads with over $100000$ comments and an average of $12$ turns per thread.\\
\textit{Open-Subtitles}: Additionally, we show results with the unannotated Open-Subtitles dataset \citep{opus_dataset} (we randomly sample up to $2$ million dialogs for training and validation). We tag the dataset with dialog attributes using pre-trained classifiers.

We experiment with two types of dialog attributes in this paper - \textit{sentiment and dialog-acts}. We annotate the utterances with sentiment tags - \textit{positive, negative, neutral} using the Stanford Core-NLP tool \citep{core-nlp}. We adopt the dialog-acts from two annotated dialog corpus - Switchboard \citep{switchboard} and Frames \citep{frames}. 

\textit{Switchboard}: Switchboard corpus\citep{switchboard} is a collection of 1155 chit-chat style telephonic conversations based on 70 topics. 
\citet{jurafsky_switchboard} revised the original tags to 42 dialog-acts. 
In our experiments, we restrict dialog-acts to the top-10 most frequently annotated tags in the corpus - \textit{Statement-non-opinion, Acknowledge , Statement-opinion, Agree/Accept, Abandoned or Turn-Exit, Appreciation, Yes-No-Question, Non-verbal, Yes answers, Conventional-closing}. We consider the top-10 frequently annotated tags as a simple solution to avoid the class imbalance issue (the \textit{Statement-non-opinion} act is tagged 72824 times, while \textit{Thanking} is tagged only 67 times) for training the dialog attribute classifiers.

\textit{Frames}: Frames\citep{frames} is a task oriented dialog corpus collected in the \textit{Wizard-of-Oz} fashion.
It comprises of 1369 human-human dialogues with an average of 15 turns per dialog. 
The wizards had access to a database of hotels and flights information and had to converse with users to help finalize vacation plans.
The dataset has 20 different types of dialog-acts annotations. 
Like the Switchboard corpus, we adopt the top 10 frequently occurring acts in the dataset for our experiments - \textit{inform, offer, request, suggest, switch-frame, no result, thank you, sorry, greeting, affirm}.

\textbf{Model Details}: We use two-layer GRUs \citep{GRU} for both encoder and decoders with hidden sizes of 512. 
We restrict the vocabulary for both the datasets to top $25000$ frequency occurring tokens. 
The dialog attribute classifier for dialog attributes is a simple 2-layer MLP with layer sizes of $256$, and $10$ respectively.
We use the rectified linear unit (ReLU) as the non-linear activation function for the MLPs and use dropout rate of $0.3$ for the token embeddings, hidden-hidden transition matrices of the encoder and decoder GRUs.

\textbf{Training Details}: We ran our experiments in Nvidia Tesla-K80 GPUs and optimized using the ADAM optimizer with the default hyper-parameters used in \citep{salasforce_lm1,salesforce_lm2}. 
All models are trained with batch size $128$ and a learning rate $0.0001$.

\section{Experimental Results}
In this section, we present the experimental results along with qualitative analysis.

In Section \ref{sec_Dialog_Attribute_Prediction}, we discuss the dialog attribute classification results for different model architectures trained on the Reddit, Switchboard and Frames datasets.

In Section \ref{sec:Utterance_Evaluation }, we first demonstrate quantitative improvements (token perplexity/embedding based metrics) for the Attribute conditional HRED model 
with the manually annotated Reddit dataset. Further, we discuss the model perplexity improvements along with sample conversations and human evaluation results on the Open-Subtitles dataset. We annotate it with sentiment and dialog-acts (from Switchboard/Frames datasets) using pre-trained classifiers described in Section \ref{sec_Dialog_Attribute_Prediction}.

Finally, in Section \ref{sec: RL_results}, we analyze the quality of the generated responses after RL fine-tuning using \textit{diversity} scores (\textit{distinct-1},  \textit{distinct-2}), sample conversations and human evaluation results for diversity and relevance.

\subsection{Dialog Attribute Prediction} \label{sec_Dialog_Attribute_Prediction}
In this section, we present the experiments with the model architectures for the dialog attribute prediction - dialog-acts from Reddit, Switchboard and Frames datasets.
First, we demonstrate the performance of the dialog-acts classifiers on the Reddit dataset as shown in Table \ref{table:da_predict}.

\begin{table}[htbp]
\centering
 \begin{tabular}{ ccc }
 \hline
 Model & Acc(\%)\\ [0.5ex] 
 \hline
 $\textrm{F}(\mathrm{U_{t}})$ & 57  \\ 
 $\textrm{F}(\mathrm{DA_{t-1,t-2}})$ & 54    \\
 $\textrm{F}(\mathrm{U_{t}}, \mathrm{DA_{t-1,t-2}})$ & \textbf{68}  \\ [1ex] 
 \hline
\end{tabular}
\caption{Dialog-acts prediction accuracy in Reddit validation set.}
\label{table:da_predict}
\end{table}

The model $\textrm{F}(\mathrm{U_{t}})$ refers to the architecture which predicts the dialog-acts based on current utterance $\mathrm{U_{t}}$ alone. 
The tokens in the current utterance $\mathrm{U_{t}}$ are fed through a two-layer GRU and the final hidden state is used to predict the dialog-acts. 
The model $\textrm{F}(\mathrm{DA_{t-1,t-2}})$ predicts the current utterance's dialog-acts $\mathrm{DA_{t}}$ based on the dialog-acts corresponding to the previous two utterances. 
We consider the dialog-acts prediction problem as a sequence modelling problem where we feed the dialog-acts into a single-layer GRU and predict the current dialog-acts conditioned on the previous dialog-acts. We settled on conditioning on the dialog-acts corresponding to the previous two utterances alone as we didn't observe any boost in the classifier performance from the older dialog-acts.
As seen in Table \ref{table:da_predict}, conditioning additionally on the dialog attributes helps improve classifier performance. 


Next, we train classifiers to predict dialog-acts of utterances of the Switchboard and Frames corpus.
In our experiments, the number of act types is 11 - the top 10 most frequently occurring acts in the corpus and "others" category covering the rest of the tags.

\begin{table}[htbp]
\centering
 \begin{tabular}{ ccc }
 \hline
 Corpus & Num Acts & Acc(\%) \\ [0.5ex] 
 \hline
 Reddit & 9 & 68.1  \\ 
 Switchboard & 11 & 67.9    \\
 Frames & 11 & 71.1\\ [1ex] 
 \hline
\end{tabular}
\caption{Dialog-acts prediction accuracy for classifiers trained on validation set of different datasets.}
\label{table:dialog-acts prediction per dataset}
\end{table}

As seen from Table \ref{table:dialog-acts prediction per dataset}, classifier performance is not really high and yet, contribute to  improvements in perplexity for the conditional Seq2Seq models (discussed in Section \ref{sec:Utterance_Evaluation }). While we aim for better classifier performance, it is important to note here that the primary objective of such dialog attribute classifiers is to tag unannotated open-domain dialog datasets. As future work, we will study how the classification errors influence response generation.

\subsection{Utterance Evaluation}\label{sec:Utterance_Evaluation }
Following \citep{hred}, we use token perplexity and embedding based metrics (average, greedy and extrema) \citep{mitchell2008vector_embedding_metrics,rus2012comparison_embedding_metrics} for utterance evaluation.

\begin{table}[htbp]
\centering
\begin{tabular}{p{1.5cm}ccc}
\hline
Metric & LM & Seq2Seq & Seq2Seq+Attr \\ [0.5ex]
\hline
Perplexity & 176 & 170 & \textbf{163} \\
Greedy & - & 0.47 & \textbf{0.54} \\
Extrema & - & 0.37 & \textbf{0.47} \\
Average & - & \textbf{0.67} & 0.62 \\
\hline
\end{tabular}
\caption{Perplexity and Embedding Metrics for the Reddit validation set.}
\label{table:metrics_for_reddit}
\end{table}

\textbf{Reddit}: First, we evaluate Seq2Seq models trained on the manually annotated Reddit corpus as shown in Table \ref{table:metrics_for_reddit}. 
\textit{Seq2Seq+Attr} refers to our model where we condition on the dialog-acts additionally. 
Please note that we use the notation "\textit{Attr}" here to maintain generality as it may refer to other dialog attributes like sentiment later in this section.
For both the baseline and conditional Seq2Seq models, we consider a dialog context involving the previous two turns as we did not observe significant performance improvement with three or more turns.
We use a 2-layer GRU language model as a baseline for comparison.
As seen from Table \ref{table:metrics_for_reddit}, \textit{Seq2Seq+Attr} fares well both in terms of perplexity and embedding metrics. 
Higher perplexity observed in the Reddit corpus could be due to the presence of several topics in the dataset (exhibits high entropy) and fewer dialogs compared to other open domain dialog datasets. 

\textbf{Open-Subtitles}: With promising results on the manually tagged Reddit corpus, we now evaluate our attribute conditional HRED model on the unannotated Open-Subtitles dataset.
We tag the Open-Subtitles dataset with the sentiment tags using the Stanford Core-NLP tool \citep{core-nlp} and dialog-acts from Frames \& Switchboard corpus using the pre-trained classifiers described in Section \ref{sec_Dialog_Attribute_Prediction}.

\begin{table}[t]
\tiny
\centering
 \begin{tabular}{ llcccc }
 \hline
 & \multicolumn{5}{c}{Num Dialogs(in Millions)} \\
 \hline
 Model & Attributes & 0.2 M & 0.5 M & 1 M & 2 M   \\ [0.5ex] 
 \hline
 Seq2seq & - & 101.63 & 80.05 & 74.78 & 67.28\\ 
 Seq2seq & Sentiment & 98.61 & 79.15 & 72.23 & 66.11\\
 Seq2seq & Switchboard & 97.03 & 77.81 & 71.51 & \textbf{64.21}\\
 Seq2seq & Frames & 96.61 & 77.41 & 72.01 & 65.33\\
  Seq2seq & Sentiment, Switchboard & 96.67 & 78.01 & 72.17 & 66.01\\
 Seq2seq & Sentiment, Frames & 96.32 & 77.61 & 72.15 & 66.13\\
 Seq2seq & Switchboard, Frames & \textbf{94.80} & \textbf{77.40} & \textbf{71.18} & 65.01\\ [1ex] 
 \hline
\end{tabular}
\caption{Validation Perplexity for the Open-Subtitles dataset.}
\label{table:OPUS_Perplexity_ablation}

\end{table}

In Table \ref{table:OPUS_Perplexity_ablation}, we compare the model perplexity when trained on varying dialog corpus size. 
In most of the cases, we observe that the conditioning with acts from both the frames and switchboard yields the lowest perplexity.
We observe that the perplexity improvement is substantial for smaller datasets which is also corroborated from the experiments with the Reddit dataset.

\textbf{Human Evaluation}: Following the human evaluation setting in \citep{jiweili2016_rl}, we randomly sample 200 input message and the generated outputs from the \textit{Seq2Seq+Attr} \& \textit{Seq2Seq} models. We present each of them to 3 judges and ask them to decide which of the two outputs is 1) relevant and 2) diverse or interesting. Ties are permitted. Results for human evaluation are shown in Table \ref{table:human evaluation_RL}. We observe that \textit{Seq2Seq+Attr} performs better than the \textit{Seq2Seq} model both in terms of \textit{diversity} and \textit{relevance}. 

\begin{table}[htbp]
\centering
 \begin{tabular}{ lccc }
  \hline
 \multicolumn{4}{c}{Seq2Seq+Attr vs Seq2Seq} \\
 \hline
 Metric & Wins(\%) & Losses(\%) & Ties(\%) \\ [0.5ex] 
 \hline
 \textit{Diversity} & \textbf{42} & 24.16 & 33.84  \\ 
 \textit{Relevance} & \textbf{40.16} & 36.83 & 23.01 \\ [2ex] 
 \hline
\end{tabular}
\label{table:human evaluation_Seq2Seq_Attr}
\caption{Human Evaluation results: \textit{Seq2Seq+Attr vs Seq2Seq}}
\end{table}

Please note that the \textit{Seq2Seq+Attr} model performs better in terms of diversity compared to the relevancy. This is in line with our expectations, as the purpose of dialog attribute annotations is to help the model focus better on less-frequent responses.

\begin{table}[t]
\centering
\small
 \begin{tabular}{ r  p{4.5cm} }
 \hline
 \hline
 Input: & i wish i was home watching tv. \\
 Seq2Seq: &  i dont know what i was thinking about\\
 Seq2Seq+Attr: &  i cant wait to see it.\\
 \hline
 Input: & He used from his charity to settle legal problems. \\
 Seq2Seq: & i have no idea what youre talking about\\
 Seq2Seq+Attr: &  i dont think he is going to be a president.\\
 \hline
 Input: & tell us how you really feel \\
 Seq2Seq: & i dont understand why\\
 Seq2Seq+Attr: & lmao i could hella picture your reaction \\ [1ex] 
\hline
\hline
\end{tabular}
\caption{Sample conversations}
\label{table:Samples for Seq2Seq+Attr and Seq2Seq}
\end{table}

Additionally, we present a few sample conversations in Table \ref{table:Samples for Seq2Seq+Attr and Seq2Seq}, where we observe that the \textit{Seq2Seq+Attr} model generates more interesting responses.

\subsection{RL For Dialog Attribute Prediction} \label{sec: RL_results}
For the RL fine-tuning, we report the diversity scores of the generated responses with the models trained on the Open-Subtitles dataset in Table \ref{table:rl_results}. The diversity scores, \textit{distinct-1} and \textit{distinct-2} are computed as the fraction of uni-grams and bi-grams in the generated responses following the previous work by  \citet{li_diversity}. 

\begin{table}[htbp]
\centering
 \begin{tabular}{ ccc }
 \hline
 Model & \textit{distinct-1} & \textit{distinct-2} \\ [0.5ex] 
 \hline
 Seq2Seq & 0.004 & 0.013\\ 
 Seq2Seq+Attr & 0.005 & 0.018\\
 RL & \textbf{0.011} & \textbf{0.033}\\ [1ex] 
 \hline
\end{tabular}
\caption{Diversity scores on the Open-Subtitles validation set after RL fine-tuning .}
\label{table:rl_results}
\end{table}

We use the model conditioned on acts from both Switchboard and Frames for the \textit{Seq2Seq+Attr} and \textit{RL} cases.
The action space for the policy in this case, covers the 10 acts from Switchboard and Frames each. 
We choose a collection of commonly occurring phrases in the Open-Subtitles dataset as the set of dull responses, $\mathbb{S}$ for the reward computation in equation \ref{reward}.
We observe that the RL fine-tuning improves over the conditional seq2seq in terms of the diversity scores.

\textbf{Human Evaluation}: As described in Section \ref{sec:Utterance_Evaluation }, we present each of the 200 randomly sampled input-response pairs of the $Seq2Seq+Attr$ and $RL$ models to 3 judges and ask to them rate each sample for \textit{diversity} and \textit{relevance}. From Table \ref{table:human evaluation_RL}, we can see that the $RL$ model significantly performs better both in terms of \textit{diversity} and \textit{relevance}.

\begin{table}[htbp]
\centering
 \begin{tabular}{ lccc }
 \hline
 \multicolumn{4}{c}{RL vs Seq2Seq+Attr} \\
 \hline
 Metric & Wins(\%) & Losses(\%) & Ties(\%) \\ [0.5ex] 
 \hline
 \textit{Diversity} & \textbf{54.66} & 28.50 & 16.84  \\ 
 \textit{Relevance} & \textbf{43.33} & 26.62 & 30.05 \\ [1ex] 
 \hline
\end{tabular}
\caption{Human Evaluation results:\textit{RL vs Seq2Seq+Attr}}
\label{table:human evaluation_RL}
\end{table}

\textbf{Qualitative Analysis}: In Table \ref{table:generic_responses_statistics}, we present the percentage of the commonly occurring generic responses from the Open-Subtitles dataset in the validation set samples corresponding to the $RL$ and $Seq2Seq + Attr$ models. We observe very low percentages of such generic responses in the samples after RL fine-tuning. It is interesting to note that RL model has successfully learned to minimize the generation of other dull responses like \textit{I would love to be , I would love to see, I dont want to} apart from expected the dull responses, $\mathbb{S}$ (used in the reward computation). At the same time, RL model has scored higher in terms of the \textit{Relevancy} metric, as seen in Table \ref{table:human evaluation_RL} which indicates that the RL fine-tuning actually explores interesting responses whilst avoiding the generic responses.   

\begin{table}[htbp]
\centering
\small
 \begin{tabular}{ lcc }
 \hline
 Generic Responses & RL(\%) & Seq2Seq + Attr(\%) \\ [0.5ex] 
 \hline
 thank you so much & 7.56 & \textbf{7.32}\\ 
 i dont understand why & \textbf{0.0} & 15.64\\
 i would love to see & \textbf{0.66} & 5.65\\
 i dont know how & \textbf{0.0} & 13.97\\
 i dont want to & \textbf{1.66} & 3.99 \\
 i dont know why & \textbf{0.0} & 3.66 \\
 i would love to be & \textbf{0.99} & 2.21 \\
 i have no idea & 4.31 & \textbf{3.33} \\
 \hline
\end{tabular}
\caption{Percentage of generic responses after RL fine-tuning.}
\label{table:generic_responses_statistics}
\end{table}

Additionally, we present a few sample conversations in Table \ref{table:samples for RL}, where we observe that the \textit{RL} model generates more diverse and relevant responses.

\begin{table}[htbp]
\centering
\small
 \begin{tabular}{ r  p{4.5cm} }
 \hline
 \hline
 Input: & i'm honestly a bit confused why no one has brought me or my books any cake \\
 Seq2Seq+Attr: & i dont think i dont think anything\\
 RL: & i cant wait to see you in the city.\\

 \hline
 Input: & ive been playing spaceship with my year old niece for the past few days \\
 Seq2Seq+Attr: & i dont even know what i was talking about.\\
 RL: & this is the best thing ive ever seen.\\
 \hline
 Input: & it makes me so happy that you like them \\
 Seq2Seq+Attr: & i dont know what i was thinking about it \\
 RL: & i was just thinking about the same thing\\ [1ex] 
 \hline
 \hline
\end{tabular}
\caption{Sample conversations}
\label{table:samples for RL}
\end{table}

\section{Related Work}

There are several works focusing on dialog-acts classification and clustering based analysis \citep{reithinger1997dialogue, liu06_ecc, dialog_act_classify, ang2005automatic, da_unsupervised, stolcke2000dialogue, da_unsupervised_act}.
\citet{conditionalVAEGen} additionally add sentiment feature to the latent variables in the VAE setting for utterance generation. 
In our work, we use dialog attributes from different sources - Switchboard and Frames corpus to model utterance generation in a more realistic setting.
As for the RL setting, existing research efforts include \cite{jiweili2016_rl, dingra_rl_2016, 2016rl_tokenJ} which formulate the token prediction as a RL policy in Seq2Seq models.
However, searching over a huge vocabulary space typically involves training with huge number of samples and careful fine-tuning of the policy optimization algorithms. Additionally, as discussed in Section \ref{sec: RL for Dialog Attribute Prediction}, it requires precautionary measures to prevent the RL algorithm from removing the linguistic aspects of the generated utterances.
In another related research work, \citet{alexa} use dialog-acts as one among their hand crafted features to select responses from an ensemble of dialog systems. They use dialog-acts in their RL policy, however their action space comprises of responses from an ensemble of dialog models. 
They include dialog-acts in their features for their distributed state representation. 

\section{Conclusion}
In this work, we address the dialog utterance generation problem by jointly modeling previous dialog context and discrete dialog attributes. 
We analyze both quantitatively (model perplexity and other embedding based metrics) and qualitatively (human evaluation, sample conversations) to validate that \textit{composed} dialog attributes help generate interesting responses. 
Further, we formulate the dialog attribute prediction problem as a reinforcement learning problem. 
We fine tune the attribute selection policy network trained with supervised learning using REINFORCE and demonstrate improvements in diversity scores compared to the Seq2Seq model. 
In the future, we plan to extend the model for additional dialog attributes like emotion, speaker persona etc. and evaluate the controllability aspect of the responses based on the dialog attributes.

\bibliography{acl2019}
\bibliographystyle{acl_natbib}
\end{document}